\documentstyle[colap,psfig]{article}

\newenvironment{bullist}
    {\begin{list}{$\bullet$}
	{\parsep 0pt \itemsep 0pt \setlength{\rightmargin}{\leftmargin}}}%
    {\end{list}}

\newcommand{\drafter}{{\sc drafter}}
\newcommand{\lline}{\noindent\rule{\textwidth}{0.25mm}}

\title{Building Knowledge Bases for the Generation of \\Software Documentation
\thanks{\hspace*{1mm} This work is partially
supported by the Engineering and Physical Sciences Research Council
({\sc epsrc}) Grant \mbox{J19221}, by {\sc bc/daad arc} Project 293,
and by the Commission of the European Union Grant \mbox{{\sc
lre}-62009}.}  }

\author{C\'ecile Paris\thanks{\hspace*{1mm} Starting this Fall,
Dr. Paris' address will
be CSIRO, Division of Information Technology, Sydney Laboratory, Building
E6B, Macquarie University Campus, North Ryde, Sydney, NSW 2113,
Australia} and Keith Vander Linden\thanks{\hspace*{1mm} Starting this Fall,
Dr. Vander Linden's address will be Dept. of Mathematics and
Computer Science, Calvin College, Grand Rapids, MI 49546, USA.}\\
	ITRI, University of Brighton\\
	Lewes Road\\
	Brighton  BN2 4AT, UK\\
	\{clp,knvl\}@itri.brighton.ac.uk\\
}

\begin{document}

\maketitle

\begin{abstract}
Automated text generation requires a underlying knowledge base from
which to generate, which is often difficult to produce.  Software
documentation is one domain in which parts of this knowledge base may
be derived automatically.  In this paper, we describe \drafter, an
authoring support tool for generating user-centred software
documentation, and in particular, we describe how parts of its
required knowledge base can be obtained automatically.


\end{abstract}

\section{Introduction}

Automated text generation is becoming an attractive technology because
it allows for the generation of text in different styles and in
different languages from a single underlying knowledge base.  The
well-known problem with the technology is that this knowledge base is
often difficult to build.  In most research generation systems, this
knowledge base is essentially built by hand.
No general solution to this problem has been proposed because each
application has its own domain specific requirements.

It is clear, however, that for text generation technology to become
viable, there must be some way to obtain at least portions of the
knowledge base automatically.
There could be a program which automatically derives the knowledge
base or perhaps the knowledge base could be built as part of manual
processes that would have to be performed anyway.  Either way, the
marginal cost of adding text generation would be greatly reduced.

In this paper, we show that software documentation is an attractive
application for multilingual text generation because it is an area in
which pre-built knowledge bases are becoming available.  This is due
in large part to the advancements in the user interface design
community which we will review first.  We then discuss the nature of
the knowledge base required for the generation of documentation and
how parts of it might be derived automatically. Finally, we illustrate
this idea using \drafter, a support tool for generating multilingual
software documentation.

\section{Background}

Researchers in user interface design have started to build tools which
produce both code and documentation.  These tools tend to be based on
a central model of the interface under development, the {\em interface
model}, a formal representation which can be used not only for code
generation but also for document generation, e.g.,
\cite{puerta94:hci,moriyon94:hci}.  
Moriyon et al \shortcite{moriyon94:hci}, for example, have used the
interface model in the generation of on-line help.  Their help
messages indicate the actions a user can perform in a particular
situation and what would result from these actions.  They report,
however, that {\em task-oriented\/} help is beyond the capabilities of
their system; task-oriented help would indicate why the user might
want to perform any of the actions that are available.

In general, however, the documentation produced by these systems is
limited in two main ways: it does not correspond to task-oriented
documentation, which is, however, what end-users require
and it is usually based on simple template
generation, which does not allow flexibility with regard to the style
of the text produced or the language that is used.  These limitations
stem, on the one hand, from the fact that interface models in general
contain {\em system-oriented\/} information (e.g., what happens when a
button is pushed) but not {\em task-oriented\/} information (e.g., why
one might want to push the button), and, on the other hand, from the
focus of the research, that is system and interface design and not
natural language generation.

In the \drafter\ project, we have attempted to address these two
issues.  We address the first by providing tools that allow technical
authors to build richer interface models.  These richer models
integrate task information into the information already available in
interface models.  This task information, which is commonly found in
task models, e.g., GOMS \cite{card83:hci}, supports the production of
user-centred documentation.  We address the second by providing more
general text generation facilities which support multiple styles and
multiple languages.

\section{Representing the users' tasks}

Early in the \drafter\ project, we conducted interviews with technical
authors (mostly software documentation specialists) in order to
understand the documentation process as it currently exists, to see if
an authoring tool would be helpful, and if so how it might be used.
We found that technical authors start the documentation process by
learning how to use the interface in question, constructing a
user-oriented mental model of the product.  They frequently have no
input other than the software itself.  The authors indicated that they
would welcome tools to help them collect the appropriate information
and create a formal representation of the resulting model.  Such a
representation would support iterative construction of the
documentation and information reuse.

Building our drafting tool, therefore, required us first to determine
how to represent the model of a task, and then to build tools for
creating and manipulating this model.  Given that the general
structure of instructional texts is hierarchical, we chose a
representation that expresses a hierarchy of goals and sub-goals.  The
representation is thus similar to the traditional structures found in
AI planning, e.g., \cite{sacerdoti77:planning}, and
also to task models used in interface design, e.g., \cite{card83:hci}.
Because user documentation frequently includes information other than
the raw actions to be performed, our representation allows authors to
include information not typically found in traditional plan
representations such as: user-oriented motivational goals, helpful
side-effects, and general comments.

\begin{figure*}[t]
\begin{center}
\strut{\psfig{figure=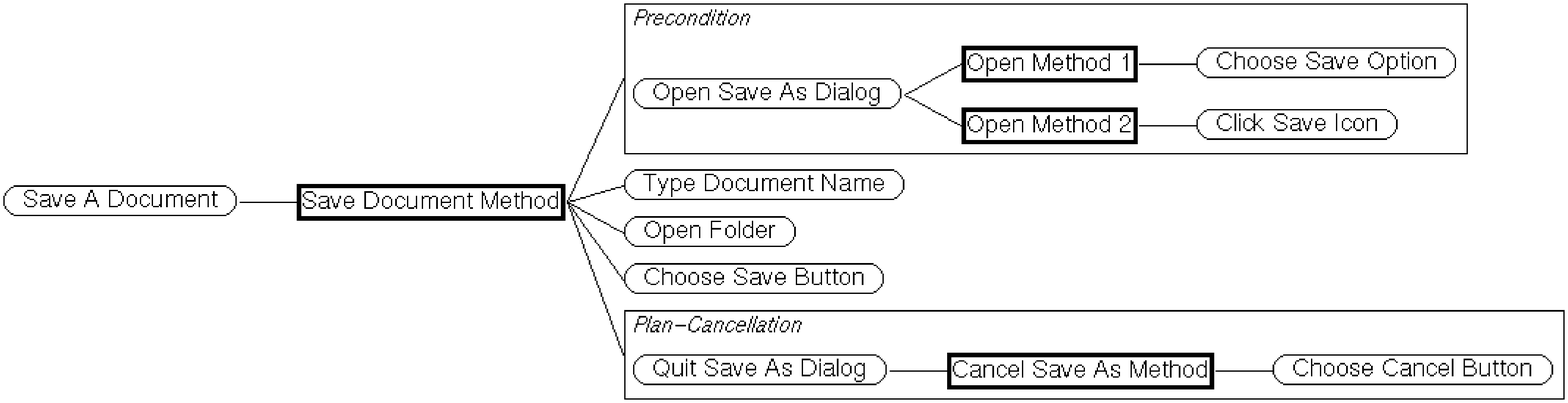,width=\textwidth}}
\caption{The Saving Procedure Graph}
\label{graph}
\end{center}
\end{figure*}

As an example, consider the representation of a sub-set of the
procedure for saving a new file in a Microsoft Word-like editor shown
in Figure~\ref{graph}.  The oval boxes in the figure represent actions
and the rectangles represent plans.  Each of the action nodes in this
structure represent interconnected complexes of procedural and
descriptive instances.  For example, the main user goal of saving a
document, represented in the figure by the action node ``Save a
Document'', is implemented in the knowledge base as a complex of
instances representing the action being performed (in this case
saving), the agent who performs action (the reader), the patient on
whom the action is performed (the current document), etc.  All of this
information is required to generate expressions of the action, but
presenting it would overly complicate the graph.

The links actually shown in the figure are based on the procedural
relations in the domain model.  For example, the plan for saving a
document (Save-Document-Plan) is linked to its goal (Save A Document),
to its precondition (Open-Save-As), and to its sub-actions of typing a
name for the current document (Type-Document-Name), opening the folder
in which it is to be saved (Open-Folder), and clicking the Save button
(Choose-Save-Button).  The precondition (Open-Save-As) must be
performed before the sub-steps may be attempted and is in turn linked
to further sub-plans (Choosing-Plan and Clicking-Plan).  This
indicates that the Save-As dialog box may be opened by either choosing
the Save option from the file menu (Choose-Save-Option) {\em or\/}
clicking the Save button on the tool bar (Click-Save-Icon).

This task model represents the procedures that a user might perform
when using an application and is the basis for generating
user-centred documentation, such as one of \drafter's texts shown in
Figure~\ref{output}.  It includes the users' high-level goals (e.g.,
``save a document'') as well as their low-level interface
manipulations (``choose the save button'').


\section{Input from the Design Process}

In our earlier work, we provided tools that supported the construction
of the task model by hand \cite{short-drafter-ijcai95}.  This went
some way to addressing the technical authors' desire for a formal
model and tools to build it.  Building the model from scratch,
however, even with the help of our menu based interface, was a tedious
and lengthy process which could potentially have rendered the
\drafter\ system impractical.  There was a clear need for facilities
to ease the input task.  In line with this, we noticed that certain
elements of the model were also present in the specifications
developed in user interface design environments.  Indeed, we found
that a number of the actions and objects in the model could be
automatically acquired from a design tool, thus providing basic
building blocks from which the full model could be constructed.

To illustrate this idea, we have built our example document editor
application in VisualWorks, a widely available interface design
environment \cite{visualworks2.0:hci}.  This tool allows one to define
the windows, dialog boxes, and other widgets relevant for the
application under development, and produces a prototype of the
interface thus specified.  Its output also includes declarative
specifications of all the widgets.  These specifications are thus
available to be exploited by other systems.  In particular, we found
that these specifications could be readily transformed into a form
appropriate for the knowledge base required by a text generation
system such as \drafter.  In our example then, we build a VisualWorks
mock-up of our word processing application, and \drafter\ derives task
model instances for all the windows and widgets in the application
(e.g., the Save-As dialog box and all its widgets) directly from the
SmallTalk source code.  \drafter\ is also able to infer the basic
interface actions that can be performed on the various interface
widgets and creates task model instances for them as well.  For
example, the system automatically defines a clicking action instance
for any ``button'' on the interface.  Similarly, it creates opening
and closing actions for all ``windows''.

Although this set of instances does not represent all the information
that could, in principle, be derived from the SmallTalk specifications
of the editor application, it nevertheless simplifies greatly the
technical author's task of knowledge specification by providing the
building blocks from which higher-level procedures can be defined.  In
the case of our admittedly simple example, seven of the nine actions
in the procedural structure are automatically specified.  The author
is required to specify only the main user goal action and the three
plan nodes.  This is, therefore, a step towards automatically building
the knowledge base required for the generation system.  It is also a
step towards integrating the design and documentation processes, which
is now widely recognised as being desirable.  In our current work, we
are investigating how more of the design knowledge can be made
accessible and understandable to the technical authors, and what other
tools would further facilitate the authors' task. We are also looking
at a tighter integration of the design and documentation processes,
one in which the individuals involved work together during design.

\section{DRAFTER}

We now describe \drafter, a technical authoring tool which supports
the construction of the task model discussed above and the drafting of
multilingual instructions from that model.  We will focus on how it
supports the author in augmenting the information automatically
acquired from the interface design tool.  \drafter's general
architecture, shown in Figure~\ref{data-flow}, is based on two main
processing modules:

\begin{figure*}[t]
\begin{center}
\strut{\psfig{figure=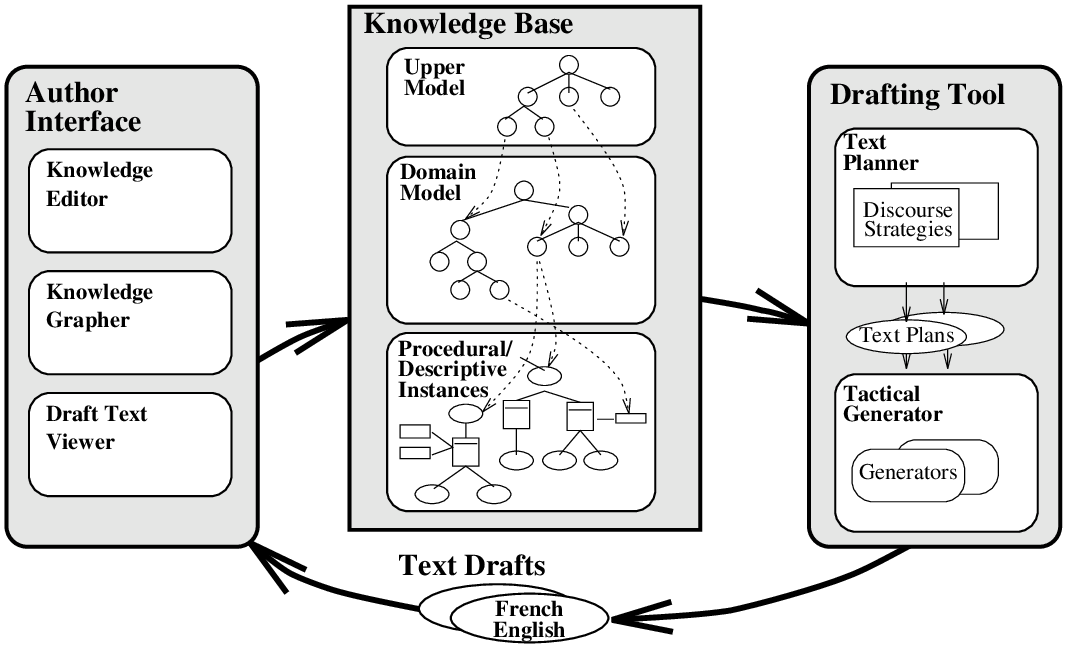,height=2.5in}}
\caption{Dataflow in \drafter}
\label{data-flow}
\end{center}
\end{figure*}

\begin{bullist}
\item The {\em Author Interface\/} (shown on the far left of the
diagram) allows authors to build a task model and to control the
drafting process.
\item The {\em Drafting Tool\/} (shown on the far right of the
diagram) comprises two major components: the Text Planner and the
Tactical Generator.  The Text Planner determines the content and
structure of the text as well as the detailed structure of the
sentences therein.  The Tactical Generator performs the surface
realisation of the sentences.
\end{bullist}

The knowledge base (in the middle of the figure) underlies the task
model built by the technical author.  The Drafting Tool takes this
representation as input and produces English and French drafts of the
appropriate tutorial instructions.  In this section we detail each of
these components in the context of an example.

\subsection{The Knowledge Base}

The knowledge base supports the construction of the task model
discussed above.  It is an hierarchical structure implemented in {\sc loom}
\cite{macgregor88:kr}.  The root is the Penman Merged
Upper Model \cite{kpml-manual}, an ontology of distinctions
relevant in expressing actions, objects, and qualities in natural
language.  The knowledge base contains further layers corresponding
to: (1) the concepts and relations general to all instructions; (2)
those general only to software interfaces; and (3) those specific to
the chosen software application domains (in our case text processing
tools).

Using the \drafter\ interface, technical authors specify hierarchical
task models, such as the one shown in Figure~\ref{graph}, by building
nodes and connecting them with the appropriate procedural relations.
The low-level building blocks of the task model are derived
automatically, and \drafter \ allows the technical author to connect
them and add higher-level task information as appropriate, using an
interface based on controlled language and the use of menus to guide
the author.

\subsection{The Interface}

\drafter's interface is implemented in {\sc clim} and includes the following
modules:

\begin{bullist}
\item The {\em Knowledge Editor} allows the author to construct and
maintain the procedural representation;
\item The {\em Knowledge Grapher} allows the author to visualise the
hierarchical structure of the procedural representation;
\item The {\em Draft Text Viewer} allows the author to view and edit
the automatically generated English and French drafts.
\end{bullist}

\noindent
These functions can be invoked
from menus or from mouse-sensitive objects in a style common to
systems such as Motif.

\subsubsection{The Knowledge Editor}

This tool makes the structure of the knowledge base on which the task
model is built more accessible to the author.  It allows the author to
perform two basic tasks: (1) specifying the action nodes appearing in the
structure and not yet derived from the interface designed tool; and
(2) linking existing nodes together with the appropriate 
plan instances and relations.  The first of these tasks is performed
using a controlled natural language interface while the second is done
with a dialog box mechanism.

\begin{figure*}[t]
\begin{center}
\strut{\psfig{figure=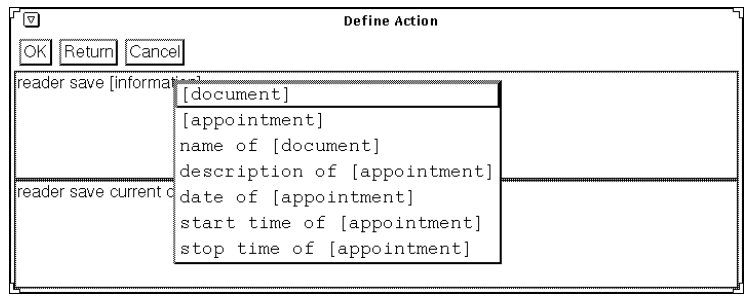,height=1.5in}}
\caption{The Controlled Natural Language Interface}
\label{cnl}
\end{center}
\end{figure*}

Specifying the nodes appearing in the task model involves specifying a
full complex of linguistic entities and role-fillers (e.g., actors,
actees, destinations).  Because this structure may include many
instances interconnected in potentially un-intuitive ways, we have
provided a Controlled Natural Language (CNL) interface for the author.
The interface is shown in Figure~\ref{cnl}.  This interface allows the
author to work in terms of sentences rather than in terms of
interconnected graphs.  The figure, for example, shows the author in
the process of specifying the node Save A Document.  The top line of
text (reader save [information]) shows the current state of the CNL
specification.  Words in brackets must be further specified.  This is
done by clicking on the word and selecting the appropriate pattern
from a list of possible expansions.  In the figure, the author has
clicked on [information] and is presented with a list of the types of
information from which [document] can be selected.  This process is
driven by a controlled natural language grammar which specifies
possible expansions at each point of the derivation.  The bottom line
of text presents a fully expanded default at each point in the
derivation.  In the figure, this CNL text is ``reader save current
document'' which could be expressed in English in a number of ways
including ``Save the current document'' and ``To save the document''.

Once the action nodes of the graph have been created, or perhaps while
they are being created, the author has the ability to link them
together using a set of predefined procedural relations: goal,
precondition, sub-action, side-effect, warning, and cancellation.
This is done with a graphical outlining mechanism.  This mechanism
allows authors to drag actions from the {\em ACTIONS} pane and drop
them on the various procedural relation slots in the workspace pane,
or, alternatively, to create new actions to fill the slots.  The
result is a procedural hierarchy such as the one shown in
Figure~\ref{graph}.


This interface allows the author to specify the procedure in several
ways.  They may start from the main goal and work down the structure,
or they may start by specifying all the low-level actions and object
and work up the structure.

\subsubsection{The Knowledge Grapher}

The Knowledge Grapher prevents the author from losing orientation by
maintaining the current state of the procedural structure in graphical
form.  This form is like that shown in Figure~\ref{graph}.  Because
the nodes are mouse-sensitive, it allows the author to initiate
construction and maintenance functions by clicking on the appropriate
nodes in the graph.  Authors can also invoke the drafting tool from
the graph.

\subsubsection{The Draft Text Viewer}

The author may draft multilingual instructions on any portion of the
procedural structure at any point in the specification process.  This
task is performed by the Drafting Tool which is briefly described in
the next section. This tool produces a draft of the instructions in
English and French.  These are presented to the author by the Draft
Text Viewer.  The presented text is mouse-sensitive, allowing the
author to access the knowledge base entry for selected part of the
text. In this way, the author can modify the underlying knowledge base
while working from the text.  In some cases the writer will decide to
modify the generated text rather than the underlying knowledge.  For
this purpose, a text editor is currently provided.

\subsection{The Drafting Tool}

When the author initiates the Drafting Tool (see
Figure~\ref{data-flow}), \drafter\ calls the Text Planner with the
discourse goal: make the user competent to perform the action
specified by the author.  The Text Planner  selects the
content appropriate for the instructions and  builds a deep
representation of the text to be generated.  This portion of the text
planning task is done by the text planner developed by Moore and Paris
\shortcite{moore-paris-cl93}.  The Text Planner then specifies the
detailed elements of the sentence structure.  This portion of the task
is done by a descendent of {\sc imagene} \cite{vanderlinden-cl95}.

Once complete, the text plans are passed to the Tactical Generator
which generates the actual text in English and French.  This task
is performed by the English and French resources of the Komet-Penman
Multi-Lingual development environment (KPML) \cite{kpml-manual}, The
drafts generated for the example procedure are shown in
Figure~\ref{output}.

\begin{figure*}[t]
\lline\\
\begin{minipage}[t]{6.5cm}
\begin{footnotesize}
{\bf To Save a Document} \\
1. Choose Save from the file menu. \\
-OR- \\
Click on the Save icon. \\
Word displays the Save As dialog box. \\
2. Type the document name in the Save Current Document As field. \\
3. Open the folder of the document. \\
4. Choose the Save button. \\
You can quit the Save As dialog box by choosing the Cancel button.
\end{footnotesize}
\end{minipage} \hspace{1cm}
\begin{minipage}[t]{7.5cm}
\begin{footnotesize}
{\bf Enregistrement d'un document} \\
1. Choisir Enregistrer dans le menu Fichier. \\
OU BIEN \\
Cliquer sur l'icone Enregistrer. \\
Word affichera la zone de dialogue Enregistrer Sous. \\
2. Introduire le titre du document dans la zone de texte Enregistrer le Document. \\
3. Ouvrir le fichier du document. \\
4. Choisir le bouton Enregistrer. \\
Vous pouvez quitter la zone de dialogue Enregistrer Sous en choisissant le bouton Annuler.
\end{footnotesize}
\end{minipage}

\medskip
\caption{Generated English and French Drafts}
\label{output}
\lline
\end{figure*}

In these texts, we see that the main user goal, that of saving a
document, is given as a title to the series of steps.  Then, the steps
to be performed to achieve this goal are given.  More detail on the
drafting process can be found elsewhere.

\section{Summary}

In this paper, we have shown that the knowledge base required to
produce user-oriented documentation automatically can be partially
obtained from user interface tools and then augmented appropriately by
technical authors.  We presented a multilingual drafting tool which
exploits output from an interface design tool and provides flexible
support to technical authors for augmenting the interface model thus
obtained in order to build the task model required to generate
documentation.  We argued that software documentation is thus an
attractive and realistic application for natural language generation.
In our current work, we are extending the percentage of the model that
can be built automatically, so as to increase the usefulness of the
system and its potential marketability.  We are also planning to
evaluate the system with technical authors.



\bibliographystyle{acl}

\begin{thebibliography}{}

\bibitem[\protect\citename{Bateman}1995]{kpml-manual}
John~A. Bateman.
\newblock 1995.
\newblock {KPML: The {\sc komet}-Penman (Multilingual) Development
  Environment}.
\newblock Technical report, Institut f{\"u}r Integrierte Publikations- und
  Informationssysteme (IPSI), GMD, Darmstadt, July.
\newblock Release 0.8.

\bibitem[\protect\citename{Card \bgroup et al.\egroup }1983]{card83:hci}
S.~K. Card, T.~P. Moran, and A.~Newell.
\newblock 1983.
\newblock {\em The Psychology of Human-Computer Interaction}.
\newblock Lawrence Earlbaum Associates, Hillsdale, NJ.

\bibitem[\protect\citename{MacGregor}1988]{macgregor88:kr}
Robert MacGregor.
\newblock 1988.
\newblock A {D}eductive {P}attern {M}atcher.
\newblock In {\em Proceedings of the 1988 Conference on Artificial
  Intelligence}, St Paul, {MN}, August. American Association of Artificial
  Intelligence.

\bibitem[\protect\citename{Moore and Paris}1993]{moore-paris-cl93}
Johanna~D. Moore and C\'ecile~L. Paris.
\newblock 1993.
\newblock Planning text for advisory dialogues: Capturing intentional and
  rhetorical information.
\newblock {\em Computational Linguistics}, 19(4):651--694.

\bibitem[\protect\citename{Moriyon \bgroup et al.\egroup }1994]{moriyon94:hci}
Roberto Moriyon, Pedro Szekely, and Robert Neches.
\newblock 1994.
\newblock Automatic generation of help from interface design models.
\newblock In {\em CHI'94 Proceedings}, Boston, Mass. Computer Human
  Interactions.

\bibitem[\protect\citename{Paris \bgroup et al.\egroup
  }1995]{short-drafter-ijcai95}
C\'ecile Paris, Keith {Vander Linden}, Markus Fischer, Anthony Hartley, Lyn
  Pemberton, Richard Power, and Donia Scott.
\newblock 1995.
\newblock A support tool for writing multilingual instructions.
\newblock In {\em IJCAI-95}, pages 1398--1404.

\bibitem[\protect\citename{Puerta and Szekely}1994]{puerta94:hci}
Angel~R. Puerta and Pedro Szekely.
\newblock 1994.
\newblock Model-based interface development.
\newblock CHI-94 Tutorial Notes.

\bibitem[\protect\citename{Sacerdoti}1977]{sacerdoti77:planning}
Earl~D. Sacerdoti.
\newblock 1977.
\newblock {\em A Structure for Plans and Behavior}.
\newblock Elsevier, New York.

\bibitem[\protect\citename{{Vander Linden} and Martin}1995]{vanderlinden-cl95}
Keith {Vander Linden} and James~H. Martin.
\newblock 1995.
\newblock Expressing local rhetorical relations in instructional text: {A}
  case-study of the purpose relation.
\newblock {\em Computational Linguistics}, 21(1):29--57, March.

\bibitem[\protect\citename{Vis}1994]{visualworks2.0:hci}
ParcPlace Systems, Inc., 999 E. Arques Avenue, Sunnyvale, CA 94086-4593, 1994.
\newblock {\em The VisualWorks Documentation}.

\end{thebibliography}

\end{document}